\documentclass[conference]{IEEEtran}
\IEEEoverridecommandlockouts
\usepackage{cite}
\usepackage{amsmath,amssymb,amsfonts}
\usepackage{algorithmic}
\usepackage{graphicx}
\usepackage{textcomp}
\usepackage{xcolor}
\usepackage{booktabs}
\def\BibTeX{{\rm B\kern-.05em{\sc i\kern-.025em b}\kern-.08em
    T\kern-.1667em\lower.7ex\hbox{E}\kern-.125emX}}

\usepackage{multirow}
\usepackage{colortbl}
\usepackage{xcolor}
\usepackage{makecell}
\usepackage{subcaption}

\usepackage{booktabs}        
\usepackage{multirow}
\usepackage[table]{xcolor}   
\definecolor{ForestGreen}{RGB}{34,139,34}
\definecolor{RowGray}{gray}{0.93}
\newcommand{\fg}[1]{\textbf{\textcolor{ForestGreen}{#1}}}

\begin{document}

\title{Vision Token Reduction via Attention-Driven Self-Compression for Efficient Multimodal Large Language Models
}

\author{
\IEEEauthorblockN{Omer Faruk Deniz, Ruiyu Mao, Ruochen Li, Yapeng Tian, Latifur Khan}
\IEEEauthorblockA{\textit{The University of Texas at Dallas}, Richardson, TX \\
\{omerfaruk.deniz, ruiyu.mao, ruochen.li, yapeng.tian, latifur.khan\}@utdallas.edu}
}

\maketitle
\newlength\savewidth\newcommand\shline{\noalign{\global\savewidth\arrayrulewidth\global\arrayrulewidth 1pt}\hline\noalign{\global\arrayrulewidth\savewidth}}
\renewcommand{\multirowsetup}{\centering}
\definecolor{ForestGreen}{RGB}{34,139,34}
\definecolor{mygray}{gray}{.85}

\begin{abstract}
Multimodal Large Language Models (MLLMs) incur significant computational cost from processing numerous vision tokens through all LLM layers. Prior pruning methods operate either before the LLM, limiting generality due to diverse encoder-projector designs or within the LLM using heuristics that are incompatible with FlashAttention. We take a different approach: rather than identifying unimportant tokens, we treat the LLM itself as the optimal guide for compression. Observing that deeper layers naturally transmit vision-to-text information, we introduce Attention-Driven Self-Compression (ADSC), a simple, broadly applicable method that progressively reduces vision tokens using only the LLM’s attention mechanism. Our method applies uniform token downsampling at selected layers, forming bottlenecks that encourage the model to reorganize and compress information into the remaining tokens. It requires no score computation, auxiliary modules, or attention modification, and remains fully compatible with FlashAttention. Applied to LLaVA-1.5, ADSC reduces FLOPs by 53.7\% and peak KV-cache memory by 56.7\%, while preserving 98.2\% of the original model performance. Across multiple benchmarks, it outperforms prior pruning approaches in both efficiency and accuracy. Crucially, under high compression ratios, our method remains robust while heuristic-based techniques degrade sharply.
\end{abstract}


\section{Introduction} MLLMs have recently emerged as powerful tools for unifying visual perception and natural language understanding \cite{llava,minigpt4}. Yet, as their capabilities scale, so do their computational demands, particularly due to the large number of vision tokens propagated through every transformer layer of the LLM \cite{fastv,tokenmerge}. Vision tokens often outnumber text tokens by an order of magnitude, dominating memory usage, increasing latency, and straining the context length of the model \cite{sparsevlm,flamingo}. This bottleneck limits the scalability and deployment efficiency of MLLMs, especially under high-resolution or long-form visual inputs.

Existing solutions often address this challenge by compressing or pruning vision tokens before they enter the LLM, typically via external modules such as Q-Formers, resamplers, or handcrafted similarity metrics \cite{blip2, recoverabletokencomp}. Although these methods can reduce computational load, they introduce architectural complexity, risk losing critical early semantic information, and are often tightly coupled to specific encoder-projector designs. More recent approaches prune tokens inside the LLM \cite{fastv, prunevid, mustdrop}, but rely on attention score-based heuristics, which require access to full attention matrices, making them incompatible with optimizations such as FlashAttention \cite{flashattn}.

In contrast to prior work that focuses on identifying and discarding unimportant tokens, we take a fundamentally different approach. We argue that the LLM itself should drive the compression process and  introduce \textbf{Attention-Driven Self-Compression (ADSC)}, a simple and general method that progressively reduces the vision token budget across transformer layers, prompting the LLM to reorganize and condense visual information as it propagates through the network. At each pruning stage, we uniformly downsample vision tokens to create an information bottleneck, encouraging the model to compress and migrate useful information into the remaining vision tokens and text tokens. This design promotes the forward flow of information from soon-to-be-dropped tokens into the survivor tokens. Due to the causal nature of self-attention, each token is further incentivized to absorb context from tokens to its left. Notably, text tokens play a central role in this process by pulling task-relevant visual information into the language stream, where it persists even after vision tokens are pruned. Through this process, the model learns to compress not by discarding seemingly low-importance tokens, but by actively reorganizing information flow, guided entirely by the LLM’s native attention dynamics.

Our approach is \textbf{intuitive, parameter-free, and broadly applicable}. It does not compute attention scores or modify the attention mechanism in any way, making it fully compatible with FlashAttention. Rather than relying on externally estimated importance, the model learns to preserve essential content through its own attention dynamics and internal reorganization, resulting in emergent compression behavior without explicit supervision.

Applied to LLaVA-1.5, our method reduces compute from 9.85 to 4.56 TFLOPs (a 53.7\% reduction) while pruning 66.7\% of vision tokens and retaining 98.2\% of the original performance. We also introduce a reverse curriculum over pruning ratios, beginning with a more aggressive compression level and gradually relaxing toward the target, which further improves compression behavior.

In summary, our contributions are as follows:
\begin{itemize}
\item We propose attention-driven self-compression, a simple and general framework that progressively prunes vision tokens \textit{inside} the LLM, treating compression as an intrinsic behavior guided by the model's attention dynamics.
\item Our method is parameter-free, broadly applicable, and fully compatible with FlashAttention, since it avoids attention-score computation and does not modify the attention mechanism.
\item We demonstrate the effectiveness of our framework through extensive experiments, showing that it substantially reduces computational cost while maintaining high accuracy, and we further show that a reverse pruning curriculum provides additional gains over alternative training strategies.
\end{itemize}
\section{Related Work}
\begin{figure*}[t]
    \centering

    \begin{subfigure}[b]{0.36\textwidth}
        \centering
        \includegraphics[width=\linewidth]{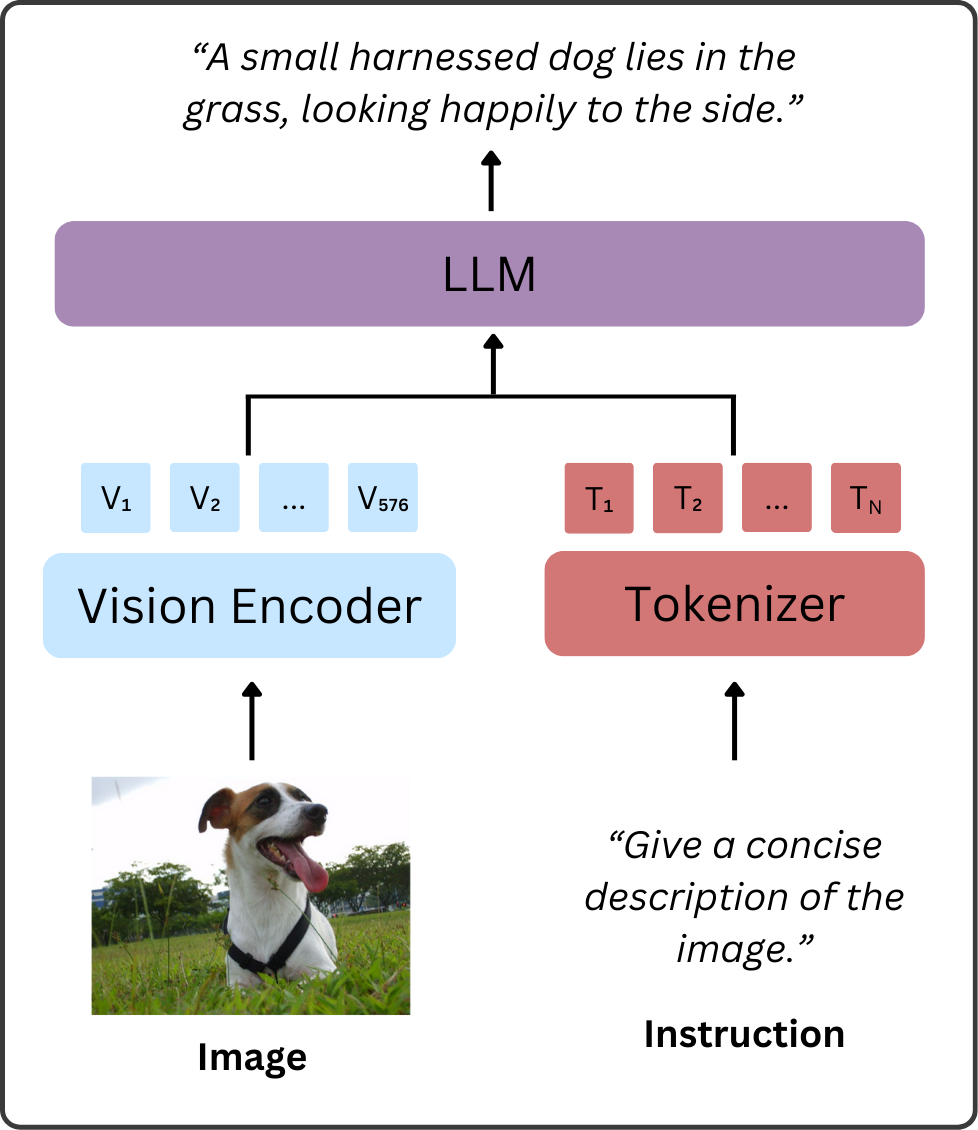}
        \caption{}
        \label{fig:pipeline-a}
    \end{subfigure}
    \hfill
    \begin{subfigure}[b]{0.627\textwidth}
        \centering
        \includegraphics[width=\linewidth]{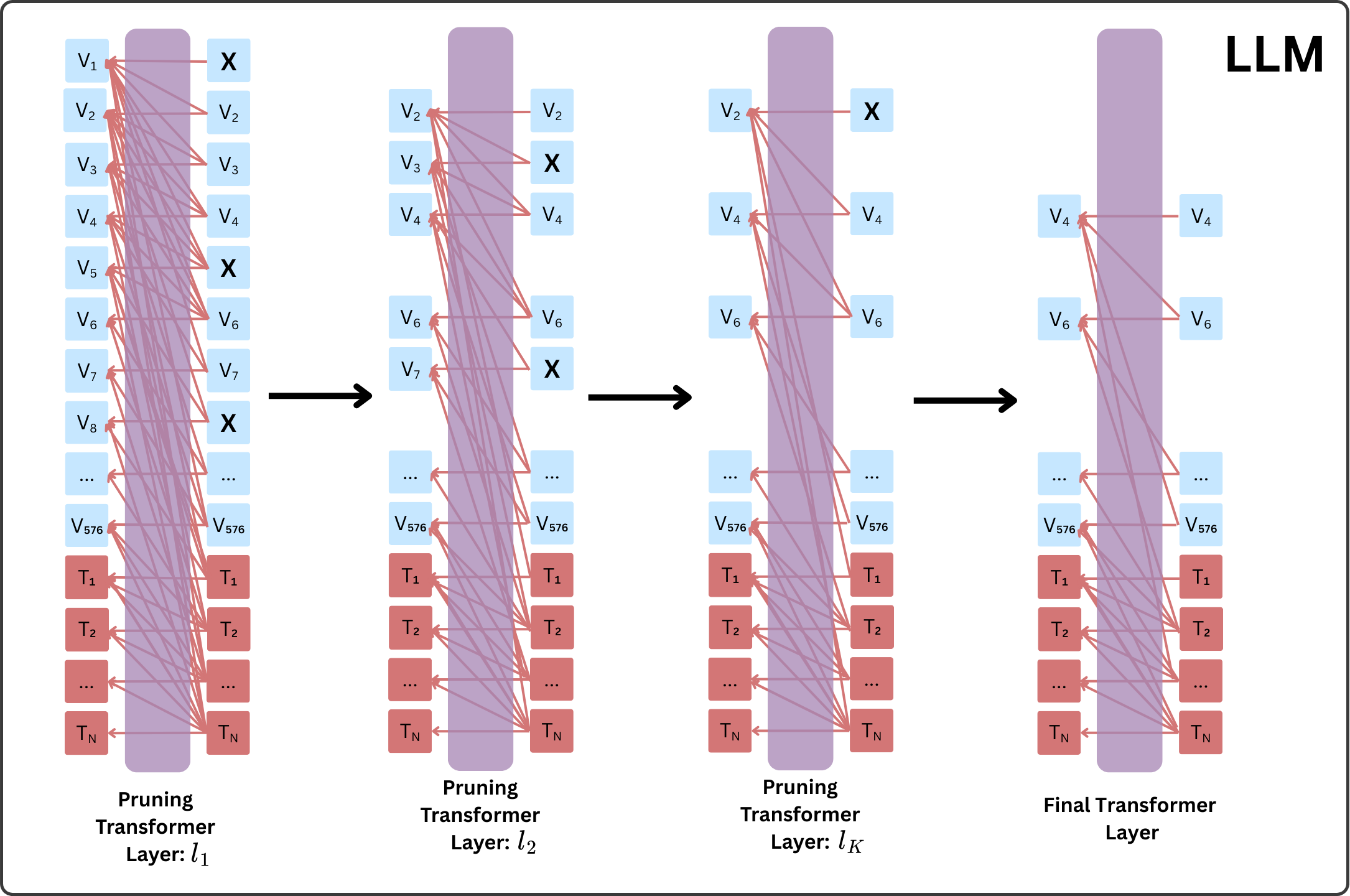}
        \caption{}
        \label{fig:pipeline-b}
    \end{subfigure}

    \caption{%
        (a) Overview of our attention-driven self-compression framework.
        (b) Vision tokens are progressively downsampled at selected LLM layers
        ($\ell_1, \ell_2, \ldots, \ell_K$), forming information bottlenecks that
        prompt the model's attention mechanism to reorganize and compress visual information.
    }
    \label{fig:pipeline}
\end{figure*}

\subsection{Multimodal Large Language Models}
Multimodal Large Language Models (MLLMs) such as GPT-4V \cite{gpt4_techreport}, LLaVA \cite{llava1.5}, and Qwen-VL \cite{qwen_vl} integrate visual and textual inputs to perform open-ended reasoning across modalities. A common design pattern in these models is to tokenize images into patch-level embeddings, often numbering in the hundreds or thousands, which are then fed into the LLM decoder. For example, LLaVA encodes a 336×336 image into 576 vision tokens, and a 672×672 image into up to 2880 tokens \cite{llava1.5}, creating substantial compute and memory overhead.

Vision tokens typically outnumber language tokens by a large margin, yet contain less structured and more spatially redundant information. While language is curated and dense, vision signals include irrelevant background and redundant details, making many vision tokens unnecessary for downstream tasks. This creates a mismatch between token volume and semantic contribution, significantly impacting inference efficiency due to the quadratic scaling of Transformer attention.

To mitigate this, prior work has explored model compression techniques such as quantization \cite{dettmers_quant8}, structured pruning \cite{llm_pruner}, and knowledge distillation \cite{llm_kd}, as well as using lightweight LLM backbones \cite{llava_phi}. However, these operate at the model level and do not directly reduce the number of vision tokens fed into the LLM, which remains a key bottleneck. Recent efforts thus turn to vision token pruning as a complementary axis of efficiency optimization.
\subsection{Token Pruning}
Token pruning has been widely studied in Vision Transformers (ViTs) to alleviate the quadratic cost of self-attention. Methods such as DynamicViT \cite{dynamicvit} and SPViT \cite{spvit} predict token importance using lightweight modules, while ToMe \cite{tome} merges semantically similar tokens through soft bipartite matching. STTS \cite{stts} extends these ideas to the spatiotemporal domain for efficient video processing. These approaches operate within the vision encoder and primarily target ViT optimization, aiming to reduce FLOPs in tasks such as image classification.

However, in MLLMs, the primary computational bottleneck lies not in the vision encoder but in the LLM. This has shifted attention toward pruning vision tokens either before or within the LLM.

\textbf{Pre-LLM Token Pruning} methods, such as LLaVA-PruMerge and related approaches \cite{prumerge, recoverabletokencomp}, compress visual inputs before they are fed into the LLM, for example by using CLS attention or token clustering on the vision-encoder outputs. In most MLLMs, this interface is implemented with model-specific encoder–projector or resampling/compression modules, so the pruning strategy becomes tightly coupled to the particular front-end design and does not transfer cleanly across architectures. Moreover, because tokens are pruned prior to entering the LLM, these methods must decide importance in the raw vision (or shallow cross-modal) space and cannot exploit the rich, contextual multimodal representations that emerge inside the LLM during reasoning.

\textbf{In-LLM Token Pruning} methods prune vision tokens \emph{inside} the decoder after vision–text fusion, using attention-derived saliency to identify redundancy. Representative methods such as FastV \cite{fastv} impose adaptive sparse attention in early layers to concentrate probability mass on a small subset of vision tokens and drop the rest downstream, while MustDrop \cite{mustdrop} performs dual-attention filtering in the prefilling stage by computing global and per-token saliency from the explicit text–vision attention matrix to select a minimal set of vision tokens to retain. However, these approaches hinge on static scores computed from the full attention matrix, a requirement incompatible with FlashAttention, which never materializes that matrix.

\textbf{Our work differs fundamentally} in that we do not rely on attention scores, handcrafted metrics, or external modules. Instead, we drop tokens at selected LLM layers, enabling the model to reorganize information flow internally. Tokens are progressively pruned based on position rather than heuristics, prompting the LLM to adaptively condense useful information into the remaining vision and language tokens. This process is fully differentiable, broadly applicable, and compatible with FlashAttention.

\section{Method}
We propose a simple yet effective method for progressively reducing vision tokens inside Multimodal Large Language Models (MLLMs) by leveraging the native transformer attention mechanism itself. Our approach, termed \textbf{Attention-Driven Self-Compression (ADSC)}, requires no architectural modification and operates entirely within the LLM.

\subsection{Preliminaries}
\textbf{FlashAttention Compatibility. }
Several prior token pruning methods rely on analyzing the attention matrix $\mathbf{A} = \text{softmax}(\mathbf{Q} \mathbf{K}^\top / \sqrt{d})$ to determine token importance. However, this design is fundamentally incompatible with efficient attention implementations such as FlashAttention~\cite{flashattn}, which do not materialize the full $N \times N$ attention matrix.
FlashAttention achieves significant speedups by computing attention in a tiled, block-wise fashion directly within high-speed SRAM, bypassing memory-intensive reads and writes to high-bandwidth memory (HBM). This not only improves wall-clock runtime but also enables fused GPU execution optimized for large-scale models.
Because FlashAttention does not expose attention weights, any method requiring them disables these optimizations, undermining both performance and scalability. This motivates a pruning approach that operates without relying on attention scores or modifying the attention computation itself.
\\
\textbf{Causal Attention. } Multimodal LLMs process input sequences of the form $\mathbf{x} = [\mathbf{x}_v; \mathbf{x}_t]$, where $\mathbf{x}_v \in \mathbf{R}^{N_v \times d}$ are vision tokens and $\mathbf{x}_t \in \mathbf{R}^{N_t \times d}$ are text tokens. These tokens propagate through $L$ decoder layers, each utilizing causal self-attention:
\begin{equation}
\text{Attn}(\mathbf{Q}, \mathbf{K}, \mathbf{V}) = \text{softmax}\left( \frac{\mathbf{Q} \mathbf{K}^\top}{\sqrt{d}} + \mathbf{M} \right)\mathbf{V},
\end{equation}
where $\mathbf{M}$ is a causal mask enforcing left-to-right dependency.
This setup ensures that each text token can attend to all prior vision tokens, enabling unidirectional information flow from vision to text. Prior work \cite{plphp} and empirical observations indicate that vision tokens are highly active in early layers but diminish in importance deeper in the model, as their information becomes absorbed by a smaller subset of remaining tokens and eventually the text tokens. This natural compression process underpins our design for progressive pruning without requiring explicit scoring.
\subsection{Vision Token Reduction via Attention-Driven Self-Compression}
To exploit the inherent redundancy of vision tokens and the natural compression dynamics of MLLMs, we propose a progressive pruning strategy that operates entirely within the LLM decoder. Rather than relying on learned importance scores or external selection modules, our method leverages the native self-attention mechanism to redistribute and condense information before dropping tokens.

We define a fixed drop ratio $r \in (0,1)$ and apply it at a predefined set of pruning layers $\mathcal{P} = \{l_1, l_2, \dots, l_K\} \subset \{1, \dots, L\}$. At each $l \in \mathcal{P}$, we retain $k^{(l)} = \lfloor (1 - r) \cdot N_v^{(l)} \rfloor$ vision tokens based on fixed, uniformly spaced spatial positions, and discard the rest:

\begin{equation}
\mathbf{x}^{(l+1)} = [\mathbf{x}_{v,\text{kept}}^{(l)}; \mathbf{x}_t],
\end{equation}

where $\mathbf{x}_t$ denotes the text tokens, which are always retained. This pruning schedule induces an exponential decay in the number of vision tokens over depth, more aggressive pruning in early layers and smaller reductions later, mirroring how visual information is gradually distilled across layers.

This design is motivated by the fundamental differences between the two modalities. Vision tokens, extracted from dense spatial grids, are often redundant and spatially correlated, making them amenable to compression. In contrast, text tokens are sequential, discrete, and information-dense; even a single dropped token may significantly degrade understanding. Moreover, text tokens carry little spatial redundancy and are harder to merge without semantic distortion. As such, we preserve the entire text sequence and restrict pruning to the vision tokens, where structured redundancy can be safely exploited.

Crucially, our method allows the model to learn how to compress rather than imposing explicit merging. Prior to each pruning layer, the self-attention mechanism operates over the full set of vision and text tokens:

\begin{equation}
\text{Attn}(\mathbf{x}_i) = \sum_{j=1}^{N} \alpha_{ij} \cdot \mathbf{V}_j,
\end{equation}
\begin{equation}
\alpha_{ij} = \frac{\exp\left( \frac{\mathbf{Q}_i \cdot \mathbf{K}_j^\top}{\sqrt{d}} \right)}{\sum_{j'=1}^{N} \exp\left( \frac{\mathbf{Q}_i \cdot \mathbf{K}_{j'}^\top}{\sqrt{d}} \right)},
\end{equation}

where each token $\mathbf{x}_i$ aggregates information from others based on attention weights $\alpha_{ij}$. Because dropped token positions are fixed across training, the model learns via positional encodings to shift and consolidate semantic content into the tokens that will be retained. Tokens destined for removal implicitly offload their information into other tokens, which in turn adapt to absorb and preserve that content. This behavior is reinforced layer by layer, allowing attention to act as a content-aware, trainable compressor.

Between pruning layers, the remaining LLM layers act as adaptive compressors: they learn to reorganize and integrate the surviving vision tokens into more abstract and compact representations. For instance, consider a scenario where 576 vision tokens are reduced to 300 at layer 8. The first 8 layers, operating on the full set, use attention to diffuse relevant content from low-salience tokens into their neighbors. After pruning, only the 300 retained vision tokens and text tokens carry forward the distilled representation, which has already absorbed contributions from the dropped tokens. This implicitly staged design divides the model into multiple compression regions. Early layers emphasize spatial aggregation and denoising, while deeper layers focus on grounding vision in language representations.

Despite using uniform token selection, the model exhibits content-aware behavior. Because the drop pattern is fixed, the model exploits its knowledge of positional priors to redirect important content toward the preserved image regions and the text tokens, which are preserved in full. Even if salient objects vary in location across images, attention enables surrounding tokens to gather and reroute their features toward stable positions. This ensures that information survives pruning, even without any explicit scoring mechanism or learned selector.

Overall, our approach treats pruning not as an external intervention, but as an internal, learnable behavior shaped by training dynamics. It leverages transformer attention to enforce a compression bottleneck, reorganizing visual content over depth, while maintaining full architectural compatibility and training stability.

\definecolor{blockgray}{gray}{0.93}

\newcommand{\blockheadergray}[1]{%
  \addlinespace[0.5ex]
  \rowcolor{blockgray}
  \multicolumn{10}{c}{\textbf{\textit{#1}}}\\
  \addlinespace[0.25ex]
}

\definecolor{HeaderGray}{gray}{0.95} 
\definecolor{OursGray}{gray}{0.80}   
\begin{table*}[t]
    \centering
    \setlength{\tabcolsep}{4pt}
    \renewcommand{\arraystretch}{1.25}
    \footnotesize
    \caption{\textbf{Main results on LLaVA-1.5-7B under different vision token budgets.}
    We report performance on six standard benchmarks and the normalized average score of each
    compressed model relative to the full 576-token LLaVA-1.5-7B baseline (right, higher is better).
    The three token budgets retain 192, 128, and 64 vision tokens on average across LLM layers, corresponding to
    66.7\%, 77.8\%, and 88.9\%
    compression. Within each budget block, the best compressed method is shown in \textbf{bold}.
    Our method, \emph{Attention-Driven Self-Compression} (ADSC), is shaded and achieves
    the highest average score and the smallest drop relative to the full model, especially
    under aggressive compression (64 tokens).}
    \label{tab:main_results}
    \vspace{2mm}
    \begin{tabular}{lcccccccccc}
        \toprule
        \multirow{2}{*}{\textbf{Method}} &
        \multirow{2}{*}{\textbf{\# Vision Tokens}} &
        \multicolumn{6}{c}{\textbf{Benchmarks} $\uparrow$} &
        \multicolumn{2}{c}{\textbf{Overall vs.\ full model}} \\
        \cmidrule(lr){3-8} \cmidrule(lr){9-10}
        & & GQA & MME & POPE & SQA & TextVQA & VQAv2 &
        Avg.\ rel.\ (\%) $\uparrow$ & Drop (pts) $\downarrow$ \\
        \midrule
        \rowcolor{HeaderGray}
\multicolumn{10}{c}{\textit{Upper bound: full LLaVA-1.5-7B (no compression)}} \\
        LLaVA-1.5-7B & 576 &
        61.9 & 1864 & 85.9 & 69.5 & 58.2 & 78.5 &
        100.0 & -- \\
        \midrule

        \rowcolor{HeaderGray}
\multicolumn{10}{c}{\textit{Retain 192 vision tokens (66.7\% compression)}} \\
        ToMe          & 192 & 54.3 & 1563 & 72.4 & 65.2 & 52.1 & 68.0 & 87.5 & 12.5 \\
        FastV         & 192 & 52.6 & 1605 & 64.8 & 69.1 & 52.5 & 67.1 & 86.4 & 13.6 \\
        PyramidDrop   & 192 & 57.3 & 1797 & 82.3 & 69.2 & 56.5 & 75.1 & 96.2 & 3.8  \\
        \rowcolor{OursGray}
        ADSC (Ours)   & 192 & \textbf{61.0} & \textbf{1801} & \textbf{85.0} & \textbf{69.5} & \textbf{56.2} & \textbf{77.1} &
                         \textbf{98.2} & \fg{1.8} \\
        \midrule

        \rowcolor{HeaderGray}
\multicolumn{10}{c}{\textit{Retain 128 vision tokens (77.8\% compression)}} \\
        ToMe          & 128 & 52.4 & 1343 & 62.8 & 59.6 & 49.1 & 63.0 & 79.6 & 20.4 \\
        FastV         & 128 & 49.6 & 1490 & 53.4 & 60.2 & 50.5 & 61.8 & 78.2 & 21.8 \\
        PyramidDrop   & 128 & 57.1 & 1761 & 82.3 & 68.4 & 56.6 & 72.9 & 95.2 & 4.8  \\
        \rowcolor{OursGray}
        ADSC (Ours)   & 128 & \textbf{60.1} & \textbf{1782} & \textbf{84.5} & \textbf{69.9} & \textbf{55.1} & \textbf{76.3} &
                         \textbf{97.4} & \fg{2.6} \\
        \midrule

        \rowcolor{HeaderGray}
\multicolumn{10}{c}{\textit{Retain 64 vision tokens (88.9\% compression)}} \\
        ToMe          &  64 & 48.6 & 1138 & 52.5 & 50.0 & 45.3 & 57.1 & 69.9 & 30.1 \\
        FastV         &  64 & 46.1 & 1256 & 38.2 & 51.1 & 47.8 & 55.0 & 67.3 & 32.7 \\
        PyramidDrop   &  64 & 47.5 & 1561 & 55.9 & 69.0 & 50.6 & 69.2 & 82.7 & 17.3 \\
        \rowcolor{OursGray}
        ADSC (Ours)   &  64 & \textbf{58.4} & \textbf{1701} & \textbf{84.0} & \textbf{69.8} & \textbf{53.2} & \textbf{73.9} &
                         \textbf{95.1} & \fg{4.9} \\
        \bottomrule
    \end{tabular}
    \vspace{-2mm}
\end{table*}

\begin{table*}[t]
\centering
\small
\setlength{\tabcolsep}{4pt}
\caption{
\textbf{Comparison of training regimes under a 192-token compression budget.}
Starting from pretrained LLaVA-1.5-7B, we evaluate full-parameter fine-tuning versus LoRA-based fine-tuning when training the ADSC compression module. Results are reported relative to the full-token (no-compression) baseline.
}
\begin{tabular}{lccccccc}
\toprule
Training Regime & GQA & MME & POPE & SQA & TextVQA & VQAv2 & Avg. rel. (\%) \\
\midrule
LLaVA-1.5-7B & 61.9 & 1864 & 85.9 & 69.5 & 58.2 & 78.5 & 100.0 \\
\midrule
ADSC + Full fine-tuning    & 60.8 & 1770 & 84.0 & 68.6 & 55.0 & 76.7 & 97.1 \\
\rowcolor{gray!10}
ADSC + LoRA fine-tuning (Ours)  & \textbf{61.0} & \textbf{1801} & \textbf{85.0} & \textbf{69.5} & \textbf{56.2} & \textbf{77.1} & \textbf{98.2} \\
\bottomrule
\end{tabular}
\label{tab:lora_vs_full}
\end{table*}

\begin{figure*}[h]
    \centering
    \includegraphics[width=\textwidth]{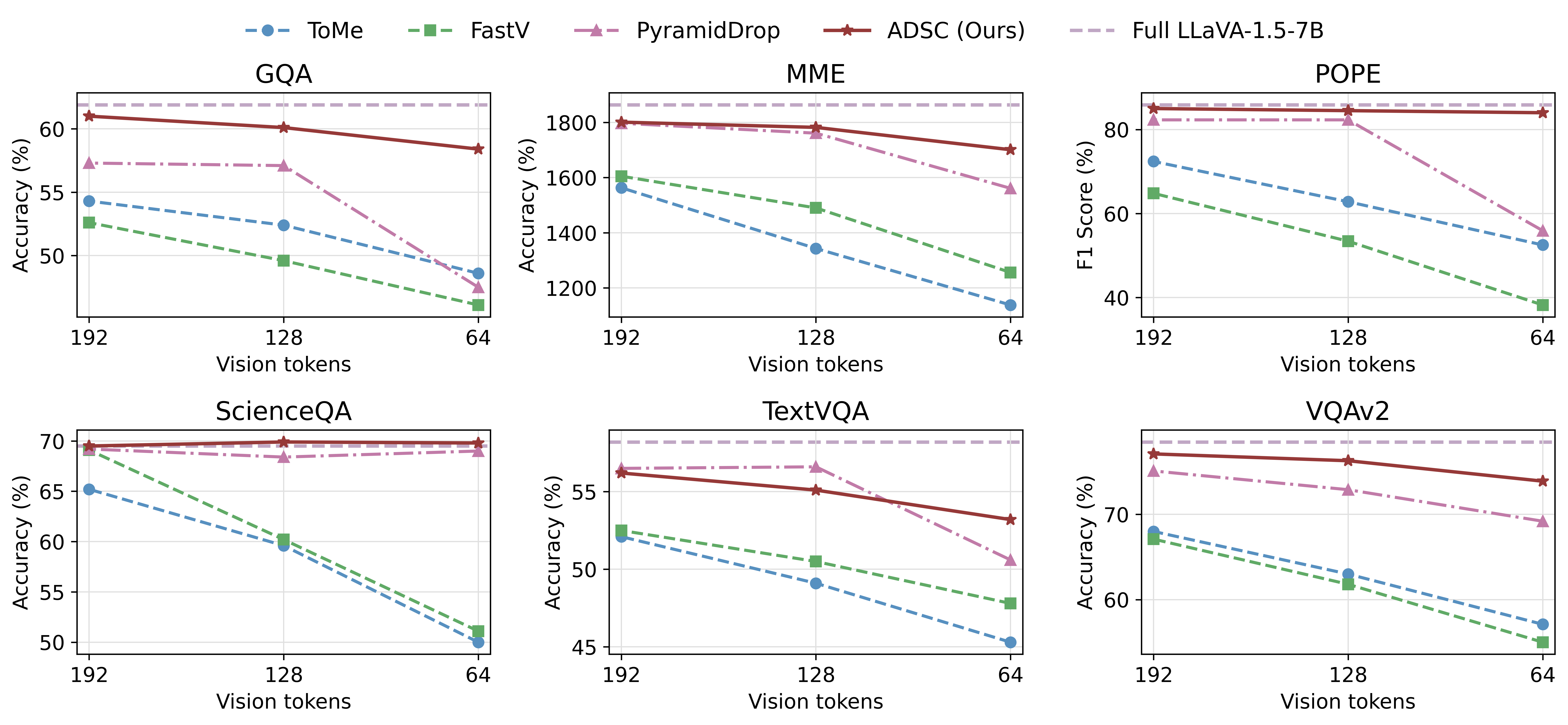}
    \caption{Performance vs. vision token count on six benchmarks (GQA, MME, POPE, ScienceQA, TextVQA, and VQAv2). Our Attention‑Driven Self‑Compression (ADSC) consistently outperforms ToMe, FastV, and PyramidDrop across all token budgets, and the performance gap widens as the number of retained vision tokens decreases. The dashed horizontal line denotes the full 576‑token LLaVA‑1.5‑7B baseline (no compression).}
    \label{fig:line_plot}
\end{figure*}

\begin{table*}[t]
\centering
\small
\setlength{\tabcolsep}{4pt}
\caption{
\textbf{Effect of curriculum strategies (192-token budget).}
All models retain the same 192-token vision budget but differ in how pruning ratios are scheduled during training. The reverse curriculum, starting with a more aggressive pruning ratio and gradually relaxing to the target, achieves the best overall performance, outperforming both direct training and the standard curriculum.
}
\begin{tabular}{lccccccc}
\toprule
Training Schedule & GQA$\uparrow$ & MME$\uparrow$ & POPE$\uparrow$ & SQA$\uparrow$ & TextVQA$\uparrow$ & VQAv2$\uparrow$ & Avg. rel. (\%)$\uparrow$ \\
\midrule
LLaVA-1.5-7B & 61.9 & 1864 & 85.9 & 69.5 & 58.2 & 78.5 & 100.0 \\
\midrule
Direct (No Curriculum)        & 60.8 & 1754 & 84.6 & 70.2 & 56.1 & 77.2 & 97.9 \\
Standard Curriculum (1.2$\times\!\rightarrow\!$1.1$\times\!\rightarrow\!$1.0$\times$) & 60.7 & 1735 & 86.3 & 70.7 & 55.9 & 77.0 & 98.1 \\
\rowcolor{gray!10}
Reverse Curriculum (0.8$\times\!\rightarrow\!$0.9$\times\!\rightarrow\!$1.0$\times$) & \textbf{61.0} & \textbf{1801} & \textbf{85.0} & \textbf{69.5} & \textbf{56.2} & \textbf{77.1} & \textbf{98.2} \\
\bottomrule
\end{tabular}
\label{tab:curriculum}
\end{table*}

\begin{table*}[t]
  \centering
  \setlength{\tabcolsep}{4pt}
  \caption{
  \textbf{Compute and memory efficiency of LLaVA-1.5-7B and ADSC at different vision-token budgets.}
  We report average forward FLOPs (prefill + decode), peak KV-cache memory per sample, and average performance
  over six benchmarks. FLOPs are measured with DeepSpeed Flops Profiler \cite{deepspeed} and KV cache with PyTorch \cite{pytorch} on a single
  NVIDIA H100 GPU.
  }
  \label{tab:efficiency_llava}
  \scriptsize
  \begin{tabular}{lccccccc}
    \toprule
    & & \multicolumn{2}{c}{\textbf{FLOPs}} 
      & \multicolumn{2}{c}{\textbf{KV Cache}} 
      & \multicolumn{2}{c}{\textbf{Avg.\ Score (6 tasks)}} \\
    \cmidrule(lr){3-4}\cmidrule(lr){5-6}\cmidrule(lr){7-8}
    \textbf{Model} 
      & \textbf{\# Vision Tokens}
      & \textbf{TFLOPs} 
      & \textbf{Rel.\ (\%)} 
      & \textbf{MB} 
      & \textbf{Rel.\ (\%)} 
      & \textbf{Score (\%)} 
      & \textbf{Rel.\ (\%)} \\
    \midrule
    LLaVA-1.5-7B  & 576 & 9.85  & 100.0 & 784.3 & 100.0 & 70.1 & 100.0 \\
    \midrule
    \multirow{3}{*}{\textbf{ADSC (Ours)}}
      & 192 & 4.56 & 46.3 & 339.5 & 43.3 & 68.9 & 98.2 \\
      & 128 & 3.71 & 37.7 & 272.0 & 34.7 & 68.3 & 97.4 \\
      &  64 & 2.85 & 29.0 & 205.2 & 26.2 & 66.7 & 95.1 \\
    \bottomrule
  \end{tabular}
\end{table*}

\subsection{Training Regime}
To optimize the model under our progressive pruning regime, we continue training from a fully trained MLLM using supervised instruction tuning (SFT) on multimodal instruction–response pairs. We adopt a parameter-efficient setup in which only a set of LoRA adapters on top of the pretrained LLaVA backbone and the projector weights are updated, while vision encoder and LLM backbone remain frozen. This allows the model to learn pruning-aware behavior while largely preserving the strong vision–language capabilities of the base model. Training is performed with the standard autoregressive cross-entropy loss over the target responses; no additional auxiliary losses are used.
\\
\textbf{Reverse curriculum over pruning ratios.}
We find that the schedule used for the pruning ratio during training has a noticeable impact on the final performance. Instead of training directly at the target pruning ratio or following a conventional curriculum that gradually increases pruning, we employ a reverse curriculum that starts from a more challenging configuration and then relaxes the pruning ratio toward the target.

Given a target pruning configuration, we first compute the corresponding budget of remaining vision tokens. Training is then divided into three phases. In the first phase, we use a more aggressive setting that allocates only 80\% of this remaining budget (i.e., 20\% fewer tokens than the final target). In the second phase, we increase the budget to 90\% of the target, and in the final phase we train with the exact target token budget. This schedule is applied consistently across all pruning layers and across different target compression regimes.

Intuitively, the initial highly compressed phase encourages the model to develop strong compression behavior under tight budgets. Subsequent phases then allow the model to refine this behavior when more vision tokens become available, leading to improved performance at the final target pruning ratio. A detailed comparison between direct training, standard curriculum, and this reverse curriculum is provided in our ablation studies.

\section{Experiments}
\subsection{Experimental Setup}
\noindent\textbf{Datasets.}
We evaluate ADSC on six public vision–language benchmarks that cover complementary capabilities: compositional reasoning, broad multimodal perception, hallucination robustness, science reasoning, and text-centric understanding. Specifically, GQA \cite{gqa} assesses compositional visual reasoning on real-world scenes. MME \cite{mme} is a comprehensive diagnostic suite with 14 perception and cognition subtasks for MLLMs. POPE \cite{pope} measures object hallucination via binary questions about object presence. ScienceQA \cite{scienceqa} requires multi-step reasoning over images, diagrams, and textual context in scientific domains. TextVQA \cite{textvqa} focuses on reading and reasoning over text embedded in natural images and VQAv2 \cite{vqav2} benchmarks general open-ended visual question answering on everyday images. For all datasets we follow the LLaVA-1.5 evaluation protocol and report the official metrics: accuracy for GQA, ScienceQA, TextVQA, VQAv2, and MME, and F1 for POPE.

\noindent
\textbf{Implementation.} We load the pretrained LLaVA-1.5-7B and train on the LLaVA-InstructionMix-665K dataset from LLaVA-1.5. During training, we use standard cross-entropy loss over the response tokens and update only the LoRA parameters and the projector, keeping the LLM backbone and vision encoder frozen. This setup enables the model to learn pruning behavior, yet it preserves the original LLaVA-1.5-7B performance much better than full fine-tuning, making it easier to retain and recover strong vision understanding capabilities under token compression. Unless specified, we have used the default hyperparameters from LLaVA-1.5 during training.

\subsection{Main Results}
Table~\ref{tab:main_results} reports results on LLaVA-1.5-7B under three average vision-token budgets (192, 128, and 64 tokens), corresponding to 66.7\%, 77.8\%, and 88.9\% vision-token compression, respectively. We compare ADSC against three strong token-reduction baselines: ToMe~\cite{tome}, FastV~\cite{fastv}, and PyramidDrop~\cite{pyramiddrop}. The rightmost columns summarize the normalized average score of each compressed model relative to the full 576-token LLaVA-1.5-7B baseline.

Under moderate compression (192 vision tokens), ADSC already operates in the near-lossless regime. It retains 98.2\% of the full-model performance, corresponding to only a 1.8-point average drop, while ToMe and FastV lose over 12 points and PyramidDrop loses 3.8 points. ADSC matches or surpasses the best baseline on five of the six benchmarks and is within 0.3 points on TextVQA, indicating that the model can absorb a 66.7\% reduction in vision tokens with negligible degradation.

At 128 tokens (77.8\% compression), ADSC remains robust, preserving 97.4\% of full performance with a 2.6-point drop. This is about 2.2 points higher than PyramidDrop (95.2\%, 4.8-point drop) and more than 17 points better than ToMe and FastV, which now lose over 20 points on average. While PyramidDrop slightly leads on TextVQA, ADSC dominates on the other benchmarks, especially GQA, MME, and POPE, suggesting that our method is particularly effective for compositional reasoning, calibration, and hallucination-sensitive tasks.

The most striking gains appear in the extreme regime of 64 tokens, where 88.9\% of vision tokens are removed. Heuristic methods degrade sharply: ToMe and FastV lose roughly 30–33 points on average, and even PyramidDrop falls to 82.7\% of full-model performance (17.3-point drop). In contrast, ADSC still retains 95.1\% of the original accuracy with only a 4.9-point drop, outperforming PyramidDrop by 12.4 points and the other baselines by more than 25 points. We hypothesize that this robustness arises because ADSC treats compression as re-encoding rather than hard selection: heuristic methods must directly pick a tiny subset of 64 tokens from a highly redundant vision grid, so any discarded token removes information that cannot be recovered and becomes increasingly harmful as the budget shrinks, whereas ADSC encourages the model to progressively migrate visual evidence into the surviving tokens and the text stream, so that the final 64 vision tokens become dense, information-rich summaries of the image rather than a sparse subset of the original grid. At this budget ADSC is the best method on all six benchmarks, including the hallucination metric POPE and OCR-style TextVQA, indicating that the model remains well-calibrated even when vision context is severely bottlenecked.

Overall, the main results demonstrate that attention-driven self-compression delivers a favorable accuracy–compression trade-off compared to the baselines. ADSC stays almost indistinguishable from the full model at 192 and 128 tokens, and remains robust even when nearly 90\% of vision tokens are removed, especially compared to heuristic-based baselines.

\subsection{Ablation Studies and Analysis}

\noindent
\textbf{Influence of Different Pruning Ratios.  }
Across all evaluated token budgets, ADSC consistently tracks the full 576-token LLaVA-1.5 baseline far more closely than heuristic pruning methods. At 192 tokens, ADSC operates in a near-lossless regime, retaining $\sim$98\% of baseline accuracy, whereas ToMe and FastV incur large drops and PyramidDrop degrades moderately. As the budget shrinks to 128 tokens, ADSC remains notably stable (97\% retention), widening the gap to heuristic approaches, which exhibit roughly linear decline, especially on GQA, MME, and POPE.

The extreme 64-token setting highlights this robustness: while ToMe and FastV collapse and PyramidDrop incurs substantial losses, ADSC still preserves $\sim$95\% of the full model’s performance and becomes the strongest method across all benchmarks. Overall, the curves show that ADSC compresses vision tokens with minimal accuracy loss, unlike heuristic pruning strategies that deteriorate sharply under high compression.

\noindent
\textbf{Influence of Training Regimes.  } 
Table~\ref{tab:lora_vs_full} shows that LoRA fine-tuning outperforms full-parameter tuning (98.2\% vs.\ 97.1\%), despite being less expressive. This arises from how each method adapts to the distribution shift introduced by token compression. Full fine-tuning updates all backbone weights, causing the model to rapidly re-optimize around compressed representations and partially overwrite its pretrained visual grounding, leading to catastrophic forgetting. 

LoRA, by contrast, constrains updates to a low-rank subspace while keeping the backbone fixed. This preserves the original multimodal representations and allows the model to learn compression-specific adjustments without erasing prior knowledge. The frozen backbone thus serves as a stable anchor, and the LoRA layers absorb the necessary adaptation. Consequently, LoRA maintains closer fidelity to the no-compression baseline, making parameter-efficient tuning especially effective for compression-aware training.

\noindent
\textbf{Influence of Curriculum Strategies.  } 
Across all three schedules in Table~\ref{tab:curriculum}, the model is evaluated with the same 192-token vision budget and only the way pruning ratios are scheduled during training differs. Under this controlled setting, the reverse curriculum consistently gives the best overall performance, achieving the highest average relative score (98.2\%) and noticeably higher MME and GQA results than both direct training and the standard curriculum.

These results suggest that making the model first operate under more aggressive compression (0.8× of the final token budget), and only later relaxing the pruning ratio toward the target (0.9× → 1.0×), is beneficial. In this regime, the model is forced early on to develop robust strategies for routing and compressing visual information under a very tight token budget; once the pruning ratio is relaxed, training primarily refines this already hard-mode compression behavior instead of learning it from scratch at an easier setting. In practice, this reverse curriculum yields better final performance than both direct training at the target ratio and the conventional curriculum that gradually increases pruning, and we therefore adopt it as our default training strategy.

\noindent
\textbf{FLOPs and KV cache.} 
Table~\ref{tab:efficiency_llava} reports compute and memory usage for LLaVA-1.5-7B and ADSC at different vision-token budgets. At a 192-token budget, ADSC reduces FLOPs from 9.85 to 4.56~TFLOPs, a 53.7\% reduction, while KV-cache memory drops from 784.3 to 339.5~MB, yet the model still preserves 98.2\% of the baseline accuracy. This indicates that more than half of both the compute and KV memory devoted to vision tokens in the original LLaVA-1.5-7B is redundant, with attention-driven self-compression, the model retains nearly all task-relevant information at roughly half the compute and memory cost. 

More aggressive budgets follow the same trend. At 128 and 64 tokens, ADSC uses only 37.7\% and 29.0\% of the baseline FLOPs, and 34.7\% and 26.2\% of the baseline KV cache, while maintaining 97.4\% and 95.1\% of the original accuracy, respectively. Even with just 64 vision tokens, less than one third of the baseline compute and roughly one quarter of the KV-cache footprint, the average score drops by only 3.4 points. Because ADSC prunes vision tokens inside the decoder, pruned tokens never produce key–value states in subsequent layers. Once a vision token is removed at a pruning stage, all deeper layers exclude it entirely, so the cached sequence shortens immediately and remains short for the rest of the forward pass. Together with the compute savings reported in Table~\ref{tab:efficiency_llava},
these improvements shift LLaVA toward a substantially more efficient operating frontier. The model preserves nearly full accuracy while operating with dramatically fewer FLOPs and a much smaller KV-cache footprint.

\section{Conclusion}
We introduced Attention-Driven Self-Compression, a simple yet effective method for reducing vision tokens inside Multimodal Large Language Models (MLLMs) by leveraging the model’s native attention dynamics. Unlike prior approaches that depend on external modules or attention score heuristics, our method imposes an information bottleneck via uniform token downsampling at selected layers, prompting the LLM itself to reorganize and condense visual information as it propagates. This strategy is fully compatible with FlashAttention, requires no architectural changes, and generalizes across many MLLM architectures. Extensive experiments demonstrate that our method consistently outperforms existing baselines, particularly under high compression, achieving substantial reductions in FLOPs and KV-cache memory with minimal loss in accuracy. Our findings suggest that token compression is not merely a filtering problem, but a learnable behavior that emerges naturally when guided by the LLM’s attention, opening new directions for efficient and scalable multimodal modeling.

\section*{Acknowledgement}
The research reported herein was supported in part by NIST under grant number 60NANB24D143, and by the National Center for Transportation Cybersecurity and Resiliency (TraCR) (a U.S. Department of Transportation National University Transportation Center) headquartered at Clemson University, Clemson, South Carolina, USA, under grant number 69A3552344812. Any opinions, findings, conclusions, and recommendations expressed in this material are those of the author(s) and do not necessarily reflect the views of NIST or TraCR. The U.S. Government assumes no liability for the contents or use thereof.

This work used the Delta and DeltaAI systems at the National Center for Supercomputing Applications through allocation CIS240609 from the Advanced Cyberinfrastructure Coordination Ecosystem: Services \& Support (ACCESS) program, which is supported by National Science Foundation grants \#2138259, \#2138286, \#2138307, \#2137603, and \#2138296.
\bibliographystyle{IEEEtran}
\bibliography{references}
\end{document}